\documentclass[conference]{IEEEtran}

\usepackage{algorithmic}
\usepackage{algorithm}
\usepackage{amsmath,amsfonts}
\usepackage{array}
\usepackage{balance}
\usepackage{booktabs} 
\usepackage[caption=false,font=normalsize,labelfont=sf,textfont=sf]{subfig}
\usepackage{cite}
\usepackage[colorlinks=false]{hyperref}
\usepackage{comment}
\usepackage{graphicx}
\usepackage{placeins} 
\usepackage{siunitx}
\usepackage{stfloats}
\usepackage{textcomp}
\usepackage{url}
\usepackage{verbatim}
\usepackage{xcolor}
\usepackage{xspace}
\usepackage{cleveref}

\definecolor{QuestionColor}{RGB}{251, 151, 39}
\definecolor{CitationColor}{RGB}{167, 204, 110}

\hypersetup{
    colorlinks=true,            %
    linkcolor=CitationColor,    %
    citecolor=CitationColor,    %
    urlcolor=gray               %
}

\crefname{section}{Sec.}{Secs.}
\Crefname{section}{Sec.}{Secs.}

\crefname{appsec}{Appendix}{Appendices}
\Crefname{appsec}{Appendix}{Appendices}

\crefname{figure}{Fig.}{Figs.}
\Crefname{figure}{Fig.}{Figs.}

\crefname{table}{Tab.}{Tabs.}
\Crefname{table}{Tab.}{Tabs.}

\makeatletter
\DeclareRobustCommand\onedot{\futurelet\@let@token\@onedot}
\def\@onedot{\ifx\@let@token.\else.\null\fi\xspace}
\makeatother

\newcommand{\eg}{e.g\onedot}
\newcommand{\ie}{i.e\onedot}

\newcommand{\Qone}{{\color{QuestionColor}\texttt{Q1}}}
\newcommand{\Qtwo}{{\color{QuestionColor}\texttt{Q2}}}
\newcommand{\Qthree}{{\color{QuestionColor}\texttt{Q3}}}
\newcommand{\Qfour}{{\color{QuestionColor}\texttt{Q4}}}

\newcommand{\Dtel}{\ensuremath{\mathcal{D}_\text{TEL}}\xspace}
\newcommand{\Dgeo}{\ensuremath{\mathcal{D}_\text{GEO}}\xspace}
\newcommand{\Daug}{\ensuremath{\mathcal{D}_\text{AUG}}\xspace}
\newcommand{\Dbase}{\ensuremath{\mathcal{D}_\text{Base}}\xspace}

\newcommand{\SeTwo}{\ensuremath{\mathrm{SE}(2)}\xspace}
\newcommand{\SoTwo}{\ensuremath{\mathrm{SO}(2)}\xspace}

\newcommand{\limo}{\textsc{LiMo}\xspace} 

\newif\ifanonymous

\begin{document}
\title{Less Is More\texorpdfstring{\kern0.12em\raisebox{-0.4ex}{\includegraphics[height=2.3ex]{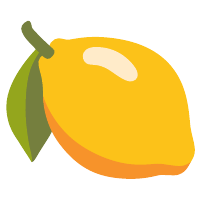}}\kern0.05em}{}: Scalable Visual Navigation from Limited Data}

\ifanonymous
\author{Author Names omitted for Anonymous Review. Paper-ID 712}
\else
\author{
    \IEEEauthorblockN{
        Yves Inglin\textsuperscript{1} \quad
        Jonas Frey\textsuperscript{2,3} \quad
        Changan Chen\textsuperscript{1} \quad
        Marco Hutter\textsuperscript{1}
    }
    \vspace{0.25em}
    \IEEEauthorblockA{
        \textsuperscript{1}ETH Zurich \quad \textsuperscript{2}Stanford University \quad \textsuperscript{3}UC Berkeley
    }
}
\fi

\maketitle

\begin{abstract}
Imitation learning provides a powerful framework for goal-conditioned visual navigation in mobile robots, enabling obstacle avoidance while respecting human preferences and social norms. 
However, its effectiveness depends critically on the quality and diversity of training data. 
In this work, we show how classical geometric planners can be leveraged to generate synthetic trajectories that complement costly human demonstrations. 
We train \emph{Less is More} (\limo), a transformer-based visual navigation policy that predicts goal-conditioned SE(2) trajectories from a single RGB observation, and find that augmenting limited expert demonstrations with planner-generated supervision yields substantial performance gains. 
Through ablations and complementary qualitative and quantitative analyses, we characterize how dataset scale and diversity affect planning performance. 
We demonstrate real-robot deployment and argue that robust visual navigation is enabled not by simply collecting more demonstrations, but by strategically curating diverse, high-quality datasets. 
Our results suggest that scalable, embodiment-specific geometric supervision is a practical path toward data-efficient visual navigation. 

\textbf{Project page:} \url{https://leggedrobotics.github.io/less-is-more}
\end{abstract}

\section{Introduction}

\begin{figure}[t]
\centering
\includegraphics[width=0.88\linewidth]{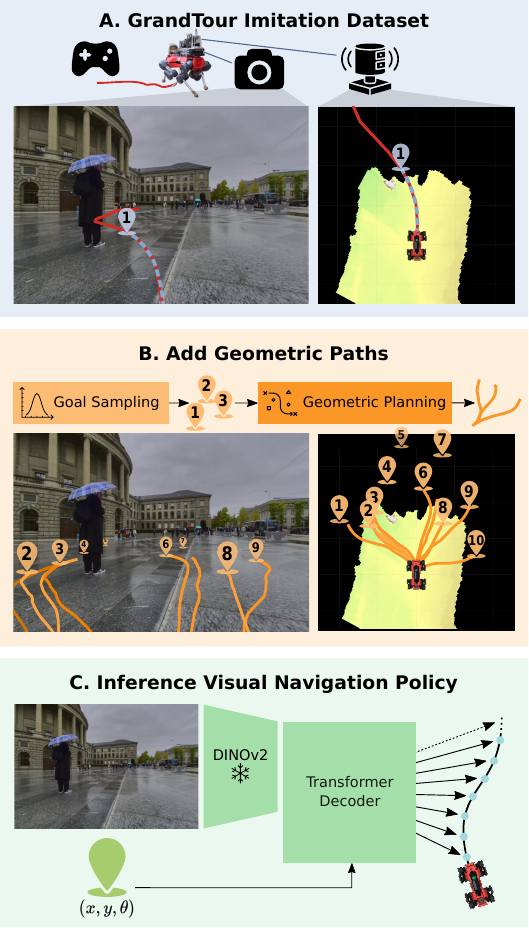}
\caption{Starting from GrandTour~\cite{frey2025boxi}, a high-quality dataset of recorded quadruped robot missions, we augment the limited real-world expert demonstrations with planner-generated trajectories to scale supervision for imitation learning. The augmented dataset is then used to train \limo, an end-to-end visual navigation policy.}
\label{fig:placeholder}
\end{figure}

\IEEEPARstart{N}{avigation} on mobile robots has traditionally followed a geometric pipeline: build a map, estimate traversability, and plan a path to the goal.
While principled and precise for collision avoidance, this approach struggles when semantic understanding or nuanced behaviors are required.
Purely geometric reasoning treats bushes or tall grass as rigid obstacles, yet it can misjudge glare, puddles, or loose gravel as safe.
Encoding social norms and complex preferences is likewise difficult in unstructured, real-world settings.

Current work focuses on learning navigation behaviors directly from RGB images using imitation learning, with an emphasis on scaling up both model capacity and dataset size by training on data from different embodiments. 
The resulting class of models, termed foundation navigation models, has shown promising preliminary results, demonstrating multi-embodiment control and goal-conditioned navigation via poses, images, or language.

Scaling-first approaches have proven particularly effective in autonomous driving, where virtually unlimited human demonstration data can be collected. 
However, in many real-world robotic settings, data is scarce, necessitating the incorporation of more first-principled approaches into the training of navigation policies. 
In such settings, transparent and rigorous analysis of model limitations becomes essential for guiding future development. 

Under this lens, we identify five core limitations that can pose significant challenges for current visual navigation models:
(i) The navigation capability relies on costly real-world expert demonstrations.
(ii) Action diversity in existing datasets is limited, biasing navigation toward simple, conservative motion.
(iii) Models must learn geometric obstacle avoidance entirely from demonstrations.
(iv) Current models trained on cross-embodiment data cannot choose actions that are desirable for a specific embodiment.
(v) Training often focuses only on feasible goals.

In response, we ask: 
How can we build upon the promising imitation learning framework while efficiently supervising navigation policies at scale to acquire both semantic understanding and robust navigation behavior, conditioned on the embodiment?

We propose \limo, an \emph{embodied visual navigation} policy that, given a single RGB image and an \SeTwo goal, predicts a sequence of waypoints for local goal-conditioned navigation.  
To enable scalable supervision, we introduce two complementary datasets derived from GrandTour~\cite{frey2025boxi}.  
The \emph{teleop} dataset transfers semantic preferences from expert trajectories, while the \emph{geometric} dataset scales supervision by sampling goals and labeling them with planner-generated trajectories.  
Because sampled goals may be infeasible, \limo also learns safe exploratory behavior and robustness to unreachable targets, making it a practical drop-in vision-based local planner.

Across held-out GrandTour missions, \limo trained on the \emph{augmented} dataset (teleop + geometric) achieves substantially higher \emph{Success
weighted by normalized inverse Path Length} (SPL)~\cite{anderson2018evaluation} than when trained solely on teleop-data, while preserving strong performance on teleop-style splits.
We show that the policy adapts paths to visible structures and selects semantically reasonable shortcuts when safe, while transparently exposing the approach's current limitations.
We further demonstrate real-robot deployment, illustrating \limo's practicality as a drop-in replacement for local path planning.
We release all datasets and pretrained models to enable reproducibility and further research. 

\textbf{Contributions.}
\begin{itemize}
    \item \limo: end-to-end, embodied, goal-conditioned navigation from a single RGB image
    \item A scalable imitation learning pipeline combining expert demonstrations and planner-generated supervision
    \item Analysis linking path-planning performance to action diversity
    \item Closed-loop real-robot deployments of \limo as a drop-in local planner
\end{itemize}

\section{Related Work}

\subsection{Geometric Path Planning}

Existing geometric planners either use raw sensory data directly or rely on aggregated spatiotemporal geometric information, such as elevation maps, to compute feasible and safe paths. 
These planners typically frame path planning as a search or optimization problem. 
Prominent approaches include graph-based search algorithms~\cite{hart1968astar} and optimization methods minimizing a cost function, such as MPPI~\cite{williams2016mppi} or iCEM~\cite{pinneri2020icem}.

These methods are interpretable and allow easy behavior tuning by adjusting the cost function, \eg, to increase clearance to obstacles or change velocity profiles.
This, together with formal guarantees of convergence to an optimal path (if it exists), makes them a favored option in industry.  
However, purely geometric sensing can miss semantically salient cues in unstructured terrain; \eg, unstable materials may appear traversable, while light grass is interpreted as a hard obstacle, missing nuanced navigation behavior. 
This motivates the use of vision to incorporate semantic understanding into robot navigation~\cite{borges2022survey, frey2023wvn}.

\subsection{Visual Navigation}
Visual navigation aims to learn navigation behaviors directly from camera input.
Previous works leverage reinforcement learning (RL)~\cite{zhu2017target} and imitation learning (IL)~\cite{loquercio2018dronet} to train visuomotor control policies.

IL has produced remarkable advances in autonomous driving (AD) ~\cite{mero2022adsurvey, liao2025diffusiondrive, nvidia2026alpamayo}.
Lately, several large-scale, high-quality AD datasets have been made available~\cite{fisher2020bdd100k, sun2020waymo}. 
In contrast, the use of IL for navigating other types of mobile robots is hindered by the limited availability of demonstration data.
Collecting expert demonstrations is costly, and existing datasets typically cover only a few hours of operation, which limits robustness and generalization.

Recent goal-image–conditioned models, such as ViNT~\cite{shah2023vint}, use transformer-based architectures to learn navigation policies intended to generalize across varied settings and robot embodiments.
NoMaD~\cite{sridhar2024nomad} extends ViNT by incorporating a diffusion head for trajectory prediction, which supports goal-conditioned navigation and unconditioned exploration. 
FlowNav~\cite{gode2025flownav} is based on NoMaD but replaces diffusion with conditional flow matching to improve inference efficiency. 
Additionally, FlowNav uses an off-the-shelf monocular depth estimation model to generate depth priors, which are then fed to the model. 
These models use a goal image (the observation when the robot is at the goal) for goal-conditioning. 
This goal image is available during training on pre-recorded robot demonstrations, but it may not be available during deployment.  

To overcome the limited data problem, \cite{shah2023vint, sridhar2024nomad, gode2025flownav} train on a mix of datasets; however, the overall scale remains limited.
The total hours reported for ViNT/NoMaD training are \SI{80}{h} across the collected datasets (see~\cref{tab:datasets}) and even less for FlowNav, which is modest given the complexity of the task.
These works compellingly demonstrate the promise of imitation learning for navigation; nevertheless, the scarcity of expert data continues to hinder generalization. 

Also notable is the significant diversity of embodiments in the training data, ranging from slow differential-drive robots operating at \SI{0.5}{m/s} to fast all-terrain vehicles reaching \SI{20}{m/s}. 
To handle this variation, the models are provided with the past five camera frames to implicitly capture embodiment, and they rely on a normalized action space where predicted waypoints are scaled by each robot’s maximum speed.

However, it is questionable whether cross-embodiment training genuinely improves model performance or whether it is primarily a compromise to increase dataset size. 
In many cases, training navigation models on embodiment-specific data may be preferable, enabling the full use of a given platform's motion capabilities. 
For instance, when training on wheeled-robot data, a quadruped robot might learn overly conservative behaviors, such as avoiding stairs, thereby unnecessarily limiting its mobility and failing to leverage its full potential.

\begin{table}
    \centering
    \caption{Overview of datasets for imitation learning}
    \label{tab:datasets}
    \begin{tabular}{@{}l l S[table-format=4.0] c c@{}}
        \toprule
        {Dataset} & {Platform} & {Total [h]} & {Env.} & {Sup.} \\
        \midrule
        \multicolumn{5}{l}{\textbf{Expert (ViNT \& NoMaD)}} \\
        BDD-100K~\cite{fisher2020bdd100k}                 & Cars        &   10 & R & H \\
        Berkeley~\cite{shah2022viking}               & Jackal      &    4 & S & H \\
        CoryHall~\cite{kahn2018coryhall}             & RC car      &    2 & H & H \\
        GoStanford~\cite{hirose2019gostanford}       & TurtleBot2  &   17 & O & H \\
        NeBula~\cite{agha2021nebula}                 & ATV         &   10 & F & H \\
        RECON~\cite{shah2021recon}                   & Jackal      &   25 & F & H \\
        SACSoN~\cite{hirose2024sacson}               & TurtleBot2  &   75 & O & H \\
        SCAND-J~\cite{karnan2022scand}               & Jackal      &    1 & W & H \\
        SCAND-S~\cite{karnan2022scand}               & Spot        &    8 & W & H \\
        Seattle~\cite{shaban2021seattle}             & Warthog     &    1 & F & H \\
        TartanDrive~\cite{triest2022tartandrive}     & ATV         &    7 & F & H \\
        \midrule
        \multicolumn{5}{l}{\textbf{Others}} \\
        LeLaN~\cite{hirose2024lelan}                     & Humans+Robots      &  129 & D & O \\
        CityWalker~\cite{liu2025citywalker}              & Humans+Cars        & 2000 & C & O \\
        FrodoBots-2k~\cite{hirose2026driveanywhere}      & Earth Rover Zero   & 2000 & U & O \\
        FrodoBots-2k(f)~\cite{hirose2026driveanywhere}   & Earth Rover Zero   &  700 & U & O \\
        OmniVLA~\cite{hirose2025omnivla}                 & Cars+Humans+Robots & 9500 & D & O \\
        TartanGround~\cite{patel2025tartanground}                 & Robots & 40 & D & P \\
        \midrule
        \multicolumn{5}{l}{\textbf{Ours}} \\
        GrandTour~\cite{frey2025boxi}    & ANYmal D    &    6 & D & H \\
        Ours                             & ANYmal D    & 2238 & D & P \\
        \bottomrule
    \end{tabular}
    \vspace{0.5em}
    \vspace{0.5em}
    \parbox{\linewidth}{
        \raggedright\footnotesize
        \textbf{Env. legend:}
        R=on-road,\;
        S=suburban,\;
        H=hallways,\;
        O=office,\;
        F=off-road,\;
        W=sidewalks,\;
        C=city walking/driving,\;
        U=urban sidewalks/ped. zones,\;
        D=highly diverse (in/out).
        \\
        \textbf{Sup.:} H=Human Expert, P=Geometric Planner, O=Other.
    }
\end{table}

LiReN~\cite{stachowicz2024liren} approaches data scarcity by continuously improving the model during deployment. 
They first use offline RL to train a navigation policy on the same multi-embodiment dataset as ViNT. 
Then they deploy the model and use an actor-critic algorithm to continuously improve it during deployment without needing a human expert to collect more data. 

LeLaN~\cite{hirose2024lelan} uses NoMaD and a vision–language model (VLM) to automatically label first-person view (FPV) YouTube videos for navigation policy learning. 

CityWalker~\cite{liu2025citywalker} also uses a dataset of FPV videos from YouTube, on which it runs visual odometry to automatically generate supervision. 
The FrodoBots-2k dataset~\cite{hirose2026driveanywhere} uses crowdsourcing; volunteers remotely drive inexpensive robots (\eg, Earth Rover Zero) to collect demonstrations in urban pedestrian zones.
This strategy scales data collection without on-site experts; however, the dataset suffers from poor tracking, inaccurate localization, and low-quality sensors.
This is addressed by using Model-Based ReAnnotation (MBRA)~\cite{hirose2026driveanywhere}, an approach that transforms a low-quality dataset into a filtered version with high-quality supervision. 

NaVILA~\cite{cheng2025navila} scales instruction-following navigation using human egocentric touring videos from YouTube. 
They estimate camera motion using an off-the-shelf model, generate language descriptions of the tours using a VLM, and train a two-stage vision–language navigation model on the resulting data. 

OmniVLA~\cite{hirose2025omnivla} is an omni-modal vision-language-action (VLA) model for navigation that supports goal conditioning via 2D goal poses, egocentric goal images, natural language instructions, and their combinations.
Built on top of OpenVLA~\cite{kim2024openvla}, OmniVLA leverages a large pretrained VLA backbone and is further trained on approximately \SI{9500}{h} of navigation data. 
This dataset is the combination of \cite{shah2023vint, hirose2024lelan, hirose2026driveanywhere, xu2017bddv} datasets from 10 different robotic platforms.
Notably, the dataset mixture is heavily skewed towards BDD-V~\cite{xu2017bddv}, \ie, on-road autonomous driving data.
They use a customized MBRA-style~\cite{hirose2026driveanywhere} approach to generate synthetic, feasible supervision for their target robots on BDD-V.

Instead of predicting waypoints directly, VENTURA~\cite{zhang2025ventura} uses an internet-pretrained image diffusion model to generate an image-space path mask conditioned on the current RGB observation and a language instruction.
Another model then converts the image-based path mask into waypoints. 
To scale supervision without manual annotations or accurate odometry, VENTURA generates path-mask labels indicating where the robot moved during deployment from egocentric videos using point tracking and adds captions using a VLM.

While \cite{hirose2024lelan, cheng2025navila, hirose2025omnivla, zhang2025ventura} study language-conditioned navigation, we focus only on goal-conditioned visual navigation. 
Despite strong progress in visuomotor policies, geometric reasoning in cluttered environments remains challenging, likely reflecting the limited geometric diversity of current datasets, which are often dominated by AD or conservative motion. 

LoGoPlanner~\cite{peng2025logoplanner} addresses this by grounding navigation in metric-scale geometry through auxiliary localization and reconstruction tasks. 
In contrast, we scale embodiment-specific geometric supervision by augmenting a high-quality legged-robot dataset with planner-generated trajectories, achieving geometry-aware performance without explicit 3D reconstruction at deployment. 
Our work extends the GrandTour~\cite{frey2025boxi} dataset's \SI{6}{h} of deployments to thousands of hours of diverse, high-quality, embodiment-specific trajectories.

\section{Method}

\begin{figure}
\centering
\includegraphics[width=0.9\linewidth]{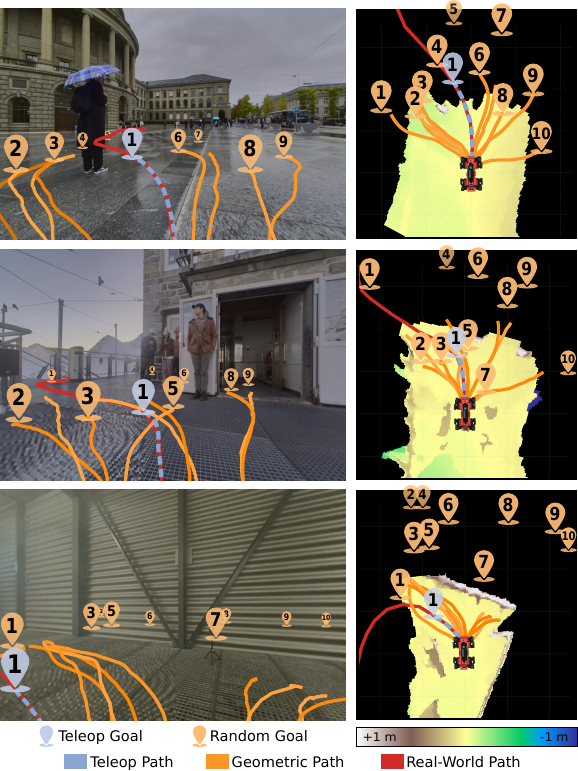}
\caption{Three representative samples from the augmented dataset \Daug. 
    Each row shows the front camera image (left) and the corresponding elevation map (right). 
    Teleoperated, geometric, and real-world paths are overlaid with their associated goals.}
\label{fig:dataset_paths}
\end{figure}

\subsection{Overview}
Given an RGB image $\mathbf{I}$ and a goal pose $\mathbf{g} = [x,y,\theta] \in \SeTwo$ in the robot-centric frame, the objective is to predict a path $\mathbf{s}_{1:N}$ of $N$ waypoints $\mathbf{s}_i=[x_i, y_i, \theta_i] \in \SeTwo$ that safely guides the robot towards the goal. 
We address this by training a policy $\pi:(\mathbf{I}, \mathbf{g})\mapsto\mathbf{s}_{1:N}$ via IL, which we call \limo.

From a base dataset \Dbase containing time-synchronized RGB video and elevation maps from expert teleoperated deployments, we extract front-camera images paired with executed \SeTwo path segments to form the \emph{teleop} dataset \Dtel (\cref{sec:method:tel-curation}).
We then sample additional goals per frame and label them with trajectories planned using a geometric MPPI planner (\cref{sec:method:mppi-planning,sec:method:geo-curation}), yielding the \emph{geometric} dataset \Dgeo.
Finally, we train \limo on the augmented dataset $\Daug=\Dtel\cup\Dgeo$ (\cref{sec:method:model}).

\subsection{MPPI Planning} 
\label{sec:method:mppi-planning}
MPPI is a gradient-free optimization algorithm used for robot motion and path planning.
Originally introduced in~\cite{williams2016mppi}, MPPI's effectiveness has been demonstrated in several real-world experiments, such as high-speed driving and whole-body control of legged robots \cite{williams2016mppi, juan2025bodymppi}.
We use MPPI to solve the following optimization problem:
\begin{equation}
    \min_{\mathbf{a}_{0:N-1}} \quad J(\mathbf{s}_0, \mathbf{a}_{0:N-1}),
\end{equation}
where $\mathbf{a}_{0:N-1}$ is a sequence of command velocities $\mathbf{a}_i = [v_i^x, v_i^y, \omega_i]$, with $v_i^x, v_i^y, \omega_i$ representing the linear and angular velocities, respectively, and 
$J:(\mathbf{s}_0, \mathbf{a}_{0:N-1})\mapsto c\in \mathbb{R}$ denoting a predefined cost function that encourages goal-reaching and safety, conditioned on the current state $\mathbf{s}_0 \in \SeTwo$.

MPPI solves the optimization by iteratively sampling a population of command velocity sequences $\mathbf{a}_{0:N-1}$.
It then uses $J$ to weight the population members and update the population distribution, preferring low-cost members. 

The future states are computed based on the current state $\mathbf{s}_0$ and a dynamics model $\mathbf{s}_{i+1} = f(\mathbf{s}_i, \mathbf{a}_i)$. 
The dynamics model $f$ can be learned or classically derived. 
In our case, we use a simple piecewise-constant velocity model in \SeTwo:
\begin{align}
    \theta_{i+1} &= \theta_i + \omega_i \,\Delta t, \\
    \begin{bmatrix}
        x_{i+1} \\
        y_{i+1}
    \end{bmatrix}
    &=
    \begin{bmatrix}
        x_i \\
        y_i
    \end{bmatrix}
    +
    \mathbf{R}(\theta_i)
    \begin{bmatrix}
        v_i^x \\
        v_i^y
    \end{bmatrix}
    \Delta t,
\end{align}
where $\mathbf{R}(\theta_i)\in\SoTwo$ is the planar rotation matrix that maps body-frame velocities to the world frame and $\Delta t$ is the control timestep.

To encourage goal reaching while ensuring robot safety, we construct the cost function as follows:
\begin{equation}
\begin{split}
    J(\mathbf{s}_0, \mathbf{a}_{0:N-1}) 
    &= w_\text{trav}C_\text{trav}(\mathbf{s}_{1:N}) 
     + w_\text{goal}C_\text{goal}(\mathbf{s}_{1:N}) \\
    &\quad + w_\text{effort}C_\text{effort}(\mathbf{a}_{0:N-1}).
\end{split}
\end{equation}

\subsubsection{Traversability Cost ($C_\text{trav}$)}
We use the convolutional neural network (CNN) from~\cite{miki2022elevation, erni2023mem} to compute a traversability map $\mathbf{T}$ from the elevation map $\mathbf{E}$.
The CNN is trained to provide an embodiment-specific traversability estimate based on the robot's capabilities. 
The traversability cost is the sum of the grid cells of $\mathbf{T}$ traversed by the robot along $\mathbf{s}_{1:N}$. 

\subsubsection{Goal distance cost ($C_\text{goal}$)}
\label{sec:method:mppi-planning:goal-dist}
By thresholding $\mathbf{T}$, we build a binary obstacle grid map $\mathbf{O}$. 
From $\mathbf{O}$, a geodesic distance field to the goal is derived~\cite{asad2022fastgeodis}.
Then $C_\text{goal}$ is the sum of the geodesic distances of each waypoint to the goal, encouraging fast goal-reaching. 
When close to the goal, the heading difference $|\theta_g - \theta_i|$ is added for goal alignment.

\subsubsection{Effort cost ($C_\text{effort}$)}   
This is the weighted sum of the linear, lateral, and angular velocities $w_\text{lin}|v^x_i| + w_\text{lat}|v^y_i| + w_\text{ang}|\omega_i|$ over the command sequence, encouraging efficient motion while discouraging unnecessary sideways motion and excessive turning.

\subsection{Dataset Curation}

\subsubsection{Teleoperation Dataset (\Dtel)}
\label{sec:method:tel-curation}
\Dtel consists of fixed-horizon, goal-conditioned segments extracted from the recorded teleoperated trajectories and serves to transfer semantic preferences and motion characteristics to the policy.

\subsubsection{Geometric Dataset (\Dgeo)} 
\label{sec:method:geo-curation}
To increase the diversity and scale of the training data, we employ the previously introduced MPPI-planner to generate additional demonstrations.
For every camera frame $\mathbf{I}$, we independently sample $K$ random goals $\mathbf{g}_1,...,\mathbf{g}_K$ from a Gaussian distribution in the robot-centric coordinate frame (see \cref{fig:goals} in the Appendix). 
Given the matching elevation map $\mathbf{E}$ and the goals, MPPI plans the paths $\mathbf{s}_{1:N}^1,..., \mathbf{s}_{1:N}^K$; generating $K$ diverse training samples $(\mathbf{I}, \mathbf{g}_k, \mathbf{s}_{1:N}^k)$ that are saved in \Dgeo.

The MPPI planner reliably generates geometrically feasible paths by leveraging privileged information from accurate elevation mapping.
This enables the creation of demonstrations that are more diverse and exploratory than those obtained from real-world teleoperation alone. 
Even when a randomly sampled goal is infeasible, the planner still produces safe, embodiment-consistent behavior, either progressing toward the goal in a best-effort manner or exploring alternative routes while avoiding obstacles (see~\cref{fig:dataset_paths}).

\subsection{Policy Architecture}
\label{sec:method:model}

\begin{figure*}
\centering
\includegraphics[width=0.70\linewidth]{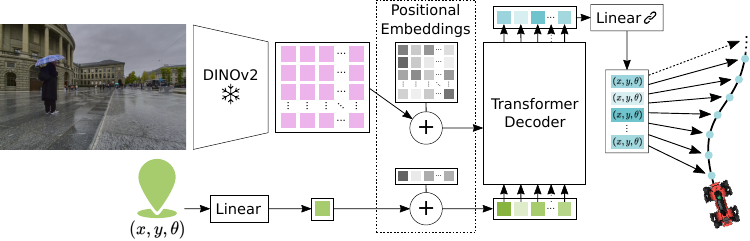}
\caption{Overview of \limo's architecture. 
    The policy takes as input a single RGB image $\mathbf{I}$, and a goal pose $\mathbf{g} = (x, y, \theta)$ in the robot-centric frame. 
    Image features are extracted using a frozen DINOv2~\cite{oquab2024dinov} encoder and combined with learned positional embeddings. 
    A transformer decoder conditioned on the goal embedding predicts $N$ waypoint embeddings, which are linearly projected to $N$ waypoints $(x, y, \theta)$, forming the output trajectory.}
\label{fig:arch}
\end{figure*}

$\pi$ receives a single RGB image $\mathbf{I}$ together with a \SeTwo goal $\mathbf{g}$ in the robot-centric coordinate frame.
It outputs $N$ waypoints in \SeTwo, also in the robot-centric frame.

First, a DINOv2~\cite{oquab2024dinov} image encoder is used to extract patch embeddings from the image $\mathbf{I}$.
The goal pose $\mathbf{g}$ is projected to the same embedding dimension and replicated $N$ times. 
We call these $N$ goal embeddings \emph{waypoint queries}. 

To every patch embedding and every waypoint query, we add an individually learned absolute positional embedding. 
Next, the \emph{waypoint queries} are passed to a 4-block transformer decoder, with the patch embeddings used as keys and values. 

The decoder outputs $N$ \emph{waypoint embeddings}, which are projected to $N$ waypoints via a linear layer with weights shared across all positions.
The resulting waypoint sequence forms the predicted plan.
An overview of the architecture is shown in \cref{fig:arch}.

Further implementation details are available in \cref{app:implementation-detail}. 

\section{Experiments}

In this section, we address the following questions: 
(\Qone)~Do the diverse geometric demonstrations effectively improve navigation performance?
(\Qtwo)~Can \limo generalize to unseen environments? 
(\Qthree)~Does our method enable embodiment-specific navigation? 
(\Qfour)~Does \limo work when deployed on a real robot? 

\subsection{Datasets}
The experiments build on the \emph{GrandTour} dataset~\cite{frey2025boxi}, which comprises \SI{6}{h} of highly diverse deployments of an ANYmal~D robot across 49 sites in Switzerland, with varying weather and lighting conditions.
From these missions, we derive three datasets, in accordance with \cref{sec:method:tel-curation,sec:method:geo-curation}:
(i) the teleoperation dataset \Dtel, (ii) the geometric-planner-labeled dataset \Dgeo with $K=10$ trajectories per image, each with $N=50$ waypoints evenly spaced over $T=$ \SI{5}{s}, and finally, (iii) the augmented dataset $\Daug = \Dtel \cup \Dgeo$.
For evaluation, we reserve six GrandTour missions exclusively as test scenes (see \cref{tab:dataset_stats} in the Appendix for further details on dataset size). 

\subsection{Qualitative Performance and Emergent Behavior}

\begin{figure*}
    \centering
    \includegraphics[width=0.90\linewidth]{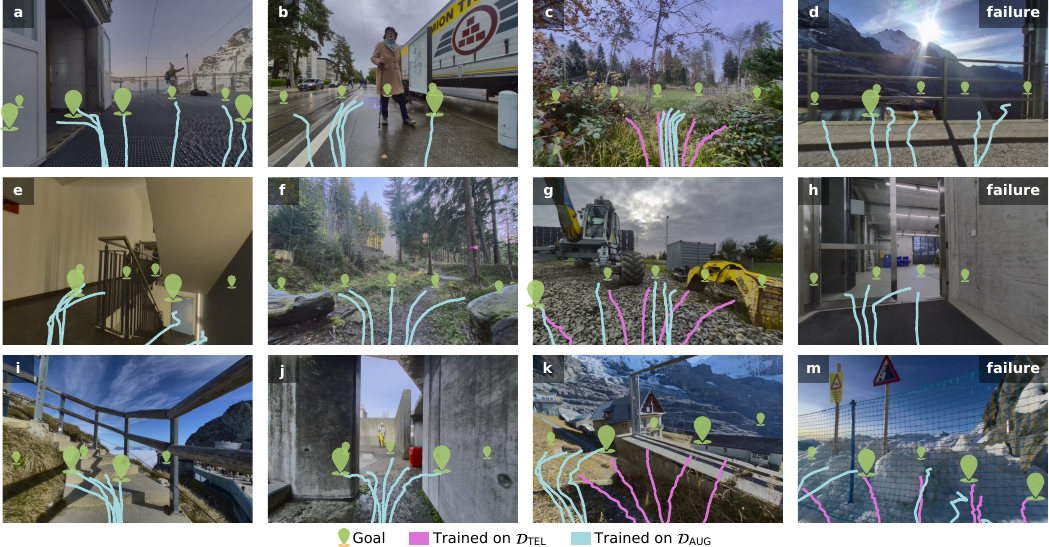}
    \caption{
        Qualitative comparison of predicted trajectories across diverse indoor and outdoor scenarios.
        Images (b, c, e, f, h, i) are included in the training set, while (a, d, g, j, k, m) are held-out scenes.
        Column three compares predictions from the \emph{teleop}-only policy with those from the \emph{augmented}-trained policy, where the latter shows much better scene understanding.
        Column four illustrates the remaining failure cases.
        \limo correctly recognizes door openings (a, j), handles stair ascent and descent consistent with ANYmal’s embodiment (e, i), and navigates unstructured natural terrain such as forests and fields (c, f).
        It also demonstrates robustness on construction sites (g) and in high-elevation outdoor environments (k), while remaining challenged by cliff edges (d), transparent surfaces, and obstacles such as glass doors and nets (h, m), and by unusual obstacles such as snow piles (m).
    }
    \label{fig:emergent}
\end{figure*}

In \cref{fig:emergent}, we present representative path predictions from \limo.
Examples (a, b, e, f, g, i, j, k) highlight a strong geometric understanding and embodiment-aware behavior: \limo identifies door openings, plans to climb and descend stairs, and navigates rough natural terrain.
Comparisons in column three (c, g, k) show that the \Daug-trained policy produces substantially more geometry-aware and obstacle-avoiding plans than the \Dtel-only baseline, clearly indicating the benefit of including geometric demonstrations~(\Qone).
The model also plans sensible trajectories in held-out missions (a, d, g, j, k), demonstrating generalization to unseen environments~(\Qtwo), and consistently chooses behaviors aligned with ANYmal’s locomotion capabilities, confirming embodiment-specific navigation~(\Qthree).

\subsection{Quantitative Results}

\subsubsection{Metrics}
Following the recommendations of \cite{anderson2018evaluation}, we evaluate trajectories on held-out missions using geodesic distances (GD) and \emph{Success weighted by normalized inverse Path Length} (SPL).
A path is deemed \emph{successful} if (i) the GD of the final waypoint to the goal is $\leq \SI{1.0}{m}$ and (ii) the robot does not collide. 
Collisions and geodesic distances are computed using traversability maps derived from the ground-truth elevation maps.
Let $S_i\in\{0,1\}$ indicate success for sample $i$, $\ell_i$ be the geodesic start-to-goal distance, and $p_i$ the path length; then the SPL is defined as:
\begin{equation}
\mathrm{SPL}=\frac{1}{M}\sum_{i=1}^{M} S_i \,\frac{\ell_i}{\max(p_i,\ell_i)}\,.
\end{equation}

\subsubsection{Baselines}

\paragraph{Geometric Planner Oracle}
The path obtained by running the planner described in \cref{sec:method:mppi-planning}. 
The geometric planner should be viewed as an approximate oracle rather than a deployable baseline. 
It plans with access to the ground-truth elevation maps that are not available to the learned policies.

Moreover, the planner is used to define the evaluation set: we consider only goals that it can reach, so as not to penalize methods for inherently infeasible targets. 

\paragraph{Straight-Line Paths}
Paths going straight from the robot to the goals, capped at \SI{5}{m} in length.

\paragraph{Real-World Paths}
The real-world paths recorded during dataset collection, only available for \Dtel goals.

\paragraph{OmniVLA (open-loop)~\cite{hirose2025omnivla}}
OmniVLA is a state-of-the-art visual navigation model initialized using the authors’ released checkpoint.
Our benchmark measures open-loop, single-shot trajectory prediction for local goal-conditioned planning, whereas OmniVLA is primarily designed for closed-loop replanning.
We therefore evaluate OmniVLA under this open-loop protocol, and the results should be interpreted accordingly.

\subsubsection{Results}

\begin{table}
  \centering
  \caption{Evaluation summary sorted by SPL}
  \label{tab:quant-results}
  \begin{tabular}{@{}l
                  S[table-format=3.1]
                  S[table-format=3.1]
                  >{\bfseries}S[table-format=3.1]@{}}
    \toprule
    {Planner} &
    {Col. (\%) $\Downarrow$} &
    {Succ. (\%) $\Uparrow$} &
    {SPL (\%) $\Uparrow$} \\
    \midrule

    \addlinespace
    \multicolumn{4}{l}{\textbf{Evaluated on \Dtel}} \\
    \textit{OmniVLA (open-loop)} & 12.6 & 67.2 & 41.5 \\
    Trained on \Dtel & 10.8 & 87.1 & 84.4 \\
    Trained on \Daug & 11.1 & 88.7 & 86.4 \\
    \textit{Straight-Line Paths} & 12.5 & 87.5 & 87.5 \\
    \textit{Real-World Paths} & 11.0 & 89.0 & 87.5 \\
    \textit{Geometric Planner} & 3.7 & 96.3 & 88.8 \\

    \addlinespace
    \multicolumn{4}{l}{\textbf{Evaluated on \Daug}} \\
    \textit{OmniVLA (open-loop)} & 14.2 & 46.5 & 33.5 \\
    Trained on \Dtel & 14.1 & 51.4 & 49.7 \\
    \textit{Straight-Line Paths} & 23.1 & 76.9 & 76.9 \\
    Trained on \Daug & 14.5 & 82.2 & 80.0 \\
    \textit{Geometric Planner} & 1.0 & 99.0 & 95.4 \\
    \bottomrule
  \end{tabular}
\end{table}

As shown in \cref{tab:quant-results}, incorporating geometric demonstrations yields substantial SPL improvements on the diverse \Daug test set, indicating that geometric supervision significantly strengthens performance in complex, unseen scenarios (\Qone, \Qtwo).

Since SPL explicitly rewards efficient, short paths, the \emph{Straight-Line} baseline performs well on \Dtel, where many goals lie roughly straight ahead, and few obstacles are encountered.
This highlights that geometric augmentation is most beneficial in diverse, obstacle-rich scenarios, where simple straight-line behavior no longer suffices.

The non-zero collision rate of the \emph{Real-World Paths} baseline mainly reflects limitations of the traversability map based collision proxy. 
Collisions are determined from elevation maps, which can label rough but physically traversable terrain as obstacles. 
As a result, safely executed recordings are sometimes counted as collisions. 
This makes the reported collision rates conservative and biased toward overestimating failures. %

We observe that \emph{OmniVLA} achieves competitive collision avoidance, even on the more challenging \Daug test set, but exhibits reduced goal-reaching performance.
This is expected in part due to the evaluation protocol: OmniVLA is trained for closed-loop replanning with fixed-length waypoint chunks rather than open-loop trajectory termination at a specified goal.

In addition, we hypothesize that training on large, cross-embodiment datasets, dominated by wheeled platforms, can further limit goal-reaching accuracy for a specific embodiment, as the model must implicitly infer platform dynamics and feasible motions solely from visual context.
Finally, the limited geometric diversity of legged-robot data within such datasets may restrict the model’s exposure to embodiment-specific maneuvers (e.g., lateral or in-place motions).
A more detailed discussion of these factors is provided in \cref{app:omnivla-comparison}.

\subsection{Ablations}

\begin{figure*}
    \centering
    \includegraphics[width=0.80\linewidth]{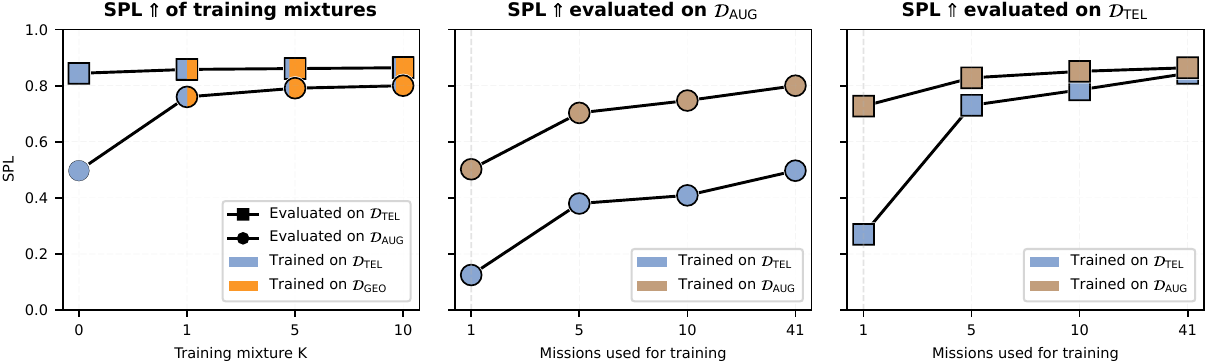}
    \caption{
        (Left) Evaluation of SPL on \Dtel and \Daug test sets trained on a varying number of augmenting geometric paths. Adding geometric samples mainly benefits performance on the more diverse \Daug test split. 
        (Middle) When evaluating goals following the \Daug distribution, policies trained solely on \Dtel consistently underperform compared to those trained on the \Daug. While training on \Daug generalizes well to \Dtel-style goals, policies trained only on \Dtel fail to generalize to unseen diverse goals with limited data.
        (Right) Evaluation of SPL on \Dtel goals trained using the \Dtel and \Daug. Training on just 5 missions of \Daug achieves comparable performance to training on \Dtel with 41 missions.
    }
    \label{fig:ablations}
\end{figure*}

\subsubsection{Training with fewer Geometric Paths per Frame}
To explore the benefits of adding geometric paths, we train policies on different mixtures of \Dtel and \Dgeo. 
The policies are evaluated on the held-out test missions with \Dtel or more diverse \Daug distributed goals (see~\cref{fig:ablations} (left)). 

\subsubsection{Training on fewer Missions}
To understand how well policies generalize when trained with fewer missions, we compare performance across varying numbers of training missions drawn from either \Dtel or \Daug. 
Policies trained on a small number of diverse \Daug missions retain strong performance on \Dtel goals, whereas policies trained only on \Dtel require substantially more missions to reach comparable SPL (see \cref{fig:ablations}). 
This highlights that diversity in the training data, rather than sheer quantity, is key for efficient generalization (\Qone, \Qtwo). 
\cref{fig:goals} in the Appendix illustrates the difference in goal diversity among datasets.

\subsection{Robot Deployment}
We deploy \limo on an ANYmal~D quadruped robot in locations outside the training scenes to demonstrate its usability in real-world closed-loop control (\Qfour), embodiment-specific behavior (\Qthree), and robustness in unseen, cluttered environments (\Qtwo).
These experiments illustrate the robot clearing obstacles, following corridors, and reacting to dynamic scene changes.

We omit on-robot comparisons to prior end-to-end visual navigation models. 
Existing approaches are designed for cross-embodiment deployment and often do not support pose-based goal conditioning. 
In contrast, \limo predicts high-resolution, goal-pose-conditioned \SeTwo trajectories that terminate at the commanded goal and are tailored to ANYmal’s embodiment.

Moreover, representative models such as OmniVLA~\cite{hirose2025omnivla} are substantially larger, cannot be executed at comparable rates on the robot’s onboard hardware, and rely on additional mechanisms such as topological maps to provide subgoals. 
We therefore deliberately choose to only highlight the real-world performance of our most powerful model.

\subsubsection{Deployment Setup}
We run \limo on a NVIDIA Jetson Orin on-board the robot in closed loop at \SI{6}{Hz}. 
We use a simple lookahead path follower node to make the robot track the predicted \SeTwo waypoints. 

\subsubsection{Incorporating Side Cameras}
Using only the front-facing camera in closed-loop deployment led to collisions, as obstacles were no longer detected once they left the forward field of view. 
Therefore, in the robot experiments, we use a multi-camera variant with additional left- and right-facing cameras, implemented via a minimal architectural extension.

More details are provided in \cref{app:side-cams}.

\subsection{Obstacle Course Navigation}

\begin{figure*}
    \centering
    \includegraphics[width=0.83\linewidth]{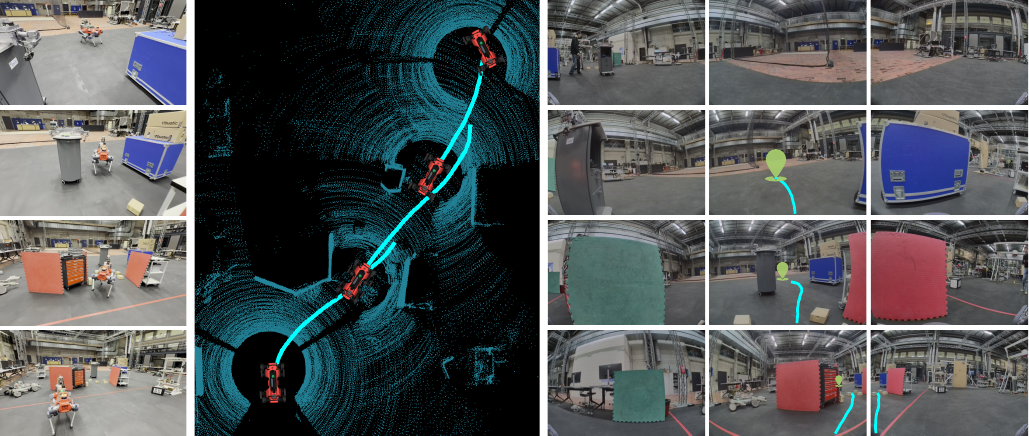}
    \caption{
    \limo navigating an indoor obstacle course shown at four different time steps.
    From left to right: (i) third-person view, (ii) bird’s-eye view of the LiDAR point cloud with predicted paths, and
    (iii) left-, front-, and right-facing camera views.
    The predicted trajectory is shown in cyan and the goal is shown in green.
    \limo generates smooth, collision-free paths while adapting to obstacles.
    The LiDAR data is recorded for visualization purposes only and is not used as input to the policy.
    }
    \label{fig:obstacle_course}
\end{figure*}

We design an indoor obstacle course comprising rigid obstacles and small traversable elements, requiring the robot to plan non-trivial paths while exploiting its legged mobility.
The robot is given a single local goal and must navigate around obstacles without collisions (see~\cref{fig:obstacle_course}).
The experiment demonstrates that \limo produces smooth, geometry-aware paths.
Despite operating purely on vision at inference time, the predicted trajectories remain consistent with the robot’s footprint and motion capabilities.

\subsection{Reacting to Dynamic Obstacles}

\begin{figure*}
    \centering
    \includegraphics[width=0.83\linewidth]{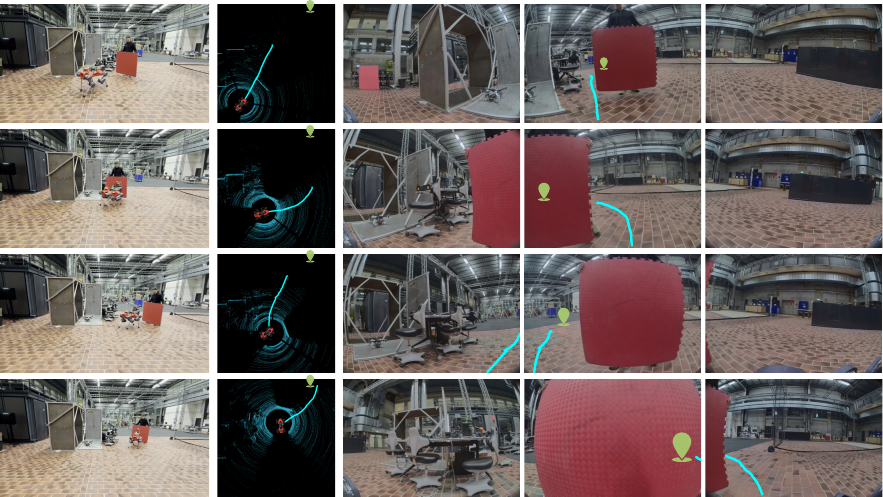}
    \caption{
    \limo reacting to a dynamic obstacle (a person carrying a red mat).
    Rows show successive time steps during execution.
    From left to right: (i) third-person view, (ii) bird’s-eye view of the LiDAR point cloud with predicted paths, and (iii) left-, front-, and right-facing camera views.
    The predicted trajectory (cyan) adapts online as the obstacle blocks the original path, leading to safe avoidance behavior.
    }
    \label{fig:dynamic_obstacle}
\end{figure*}

To test reactivity, we introduce a moving obstacle by having a person carry a large mat and intentionally block the robot’s path during execution.
\limo runs continuously at \SI{6}{Hz}, allowing the robot to replan as the obstacle moves.

As shown in \cref{fig:dynamic_obstacle}, the robot quickly deviates from its original path and selects alternative routes, including lateral and backward embodiment-specific motion, to avoid colliding.
This behavior demonstrates that \limo can handle dynamic environmental changes.

\subsection{Corridor Following}
We deploy the robot in a long corridor scenario with limited lateral clearance.
The robot must stay centered as it progresses toward a goal located to the right of the corridor exit (see~\cref{fig:corridor} in the Appendix).

\section{Limitations}
Planner supervision depends on elevation and traversability estimates and can fail in edge cases, leading to unsafe or suboptimal labels.
At inference, \limo operates on a single RGB frame without temporal memory, which limits reasoning under occlusions and causes obstacles to be forgotten once they leave the field of view.
While geometric augmentation scales supervision efficiently, visual diversity is still constrained by the underlying dataset.
\cref{fig:emergent} illustrates representative failure cases, and a continuing discussion is provided in \cref{app:limitations}.

\section{Conclusion}
We presented \limo, a goal-conditioned visual navigation policy trained on a mix of teleoperation data and scalable geometric supervision.  
By combining expert demonstrations with planner-generated trajectories that are fully embodiment-specific, we scale supervision to thousands of hours of high-quality trajectories.
Across held-out missions, geometric augmentation leads to clear improvements in SPL, especially in diverse, obstacle-rich environments.  
\limo learns geometry-aware, semantically grounded behavior aligned with ANYmal’s embodiment, and experiments on a real robot validate its utility as a practical path planner.

Overall, our results show that for vision-based goal-conditioned navigation, curating diverse, geometry-aware supervision can be more effective than relying solely on large, heterogeneous datasets.

By combining limited expert demonstrations with scalable, embodiment-specific geometric supervision, we provide a practical and data-efficient path toward robust visual navigation in real-world robotic systems.

\clearpage

\bibliographystyle{IEEEtran}
\balance
\bibliography{references}

@inproceedings{frey2025boxi,
author={Jonas Frey and Turcan Tuna and Lanke Frank Tarimo Fu and Cedric Weibel and Katharine Patterson and Benjamin Krummenacher and Matthias Müller and Julian Nubert and Maurice Fallon and Cesar Cadena and Marco Hutter},
title={{Boxi: Design Decisions in the Context of Algorithmic Performance for Robotics}},
booktitle={Proceedings of Robotics: Science and Systems (RSS)},
year={2025},
address={Los Angeles, CA, USA},
doi={10.15607/RSS.2025.XXI.134}
}

@misc{anderson2018evaluation,
title={On Evaluation of Embodied Navigation Agents}, 
author={Peter Anderson and Angel Chang and Devendra Singh Chaplot and Alexey Dosovitskiy and Saurabh Gupta and Vladlen Koltun and Jana Kosecka and Jitendra Malik and Roozbeh Mottaghi and Manolis Savva and Amir R. Zamir},
year={2018},
eprint={1807.06757},
archivePrefix={arXiv},
primaryClass={cs.AI},
url={https://arxiv.org/abs/1807.06757}
}

@article{hart1968astar,
author={Hart, Peter E. and Nilsson, Nils J. and Raphael, Bertram},
title={A Formal Basis for the Heuristic Determination of Minimum Cost Paths}, 
journal={IEEE Transactions on Systems Science and Cybernetics}, 
year={1968},
volume={4},
number={2},
pages={100-107},
doi={10.1109/TSSC.1968.300136}
}

@inproceedings{williams2016mppi,
author={Williams, Grady and Drews, Paul and Goldfain, Brian and Rehg, James M. and Theodorou, Evangelos A.},
title={Aggressive driving with model predictive path integral control}, 
booktitle={IEEE International Conference on Robotics and Automation (ICRA)}, 
year={2016},
volume={},
number={},
pages={1433-1440},
address = {Stockholm, Sweden},
doi={10.1109/ICRA.2016.7487277}
}

@inproceedings{pinneri2020icem,
author = {Pinneri, Cristina and Sawant, Shambhuraj and Blaes, Sebastian and Achterhold, Jan and Stueckler, Joerg and Rolinek, Michal and Martius, Georg},
title = {Sample-efficient Cross-Entropy Method for Real-time Planning},
booktitle = {Conference on Robot Learning (CoRL)},
year = {2020},
address = {Online}
}

@article{borges2022survey,
author={Borges, Paulo V. K. and Peynot, Thierry and Liang, Sisi and Arain, Bilal and Wildie, Matthew and Minareci, Melih G. and Lichman, Serge and Samvedi, Garima and Sa, Inkyu and Hudson, Nicolas and Milford, Michael and Moghadam, Peyman and Corke, Peter},
title={A Survey on Terrain Traversability Analysis for Autonomous Ground Vehicles: Methods, Sensors, and Challenges}, 
journal={Field Robotics}, 
year={2022},
volume={2},
pages={1567-1627},
doi={10.55417/fr.2022049}
}

@inproceedings{frey2023wvn, 
author = {Jonas Frey and Matias Mattamala and Nived Chebrolu and Cesar Cadena and Maurice Fallon and Marco Hutter}, 
title = {{Fast Traversability Estimation for Wild Visual Navigation}}, 
booktitle = {Proceedings of Robotics: Science and Systems (RSS)}, 
year = {2023}, 
address = {Daegu, Republic of Korea}, 
month = {July}, 
doi = {10.15607/RSS.2023.XIX.054} 
}

@inproceedings{zhu2017target,
author = {Zhu, Yuke and Mottaghi, Roozbeh and Kolve, Eric and Lim, Joseph J. and Gupta, Abhinav and Fei-Fei, Li and Farhadi, Ali},
title = {Target-driven visual navigation in indoor scenes using deep reinforcement learning},
booktitle = {IEEE International Conference on Robotics and Automation (ICRA)},
year = {2017},
pages = {3357–3364},
numpages = {8},
address = {Singapore, Singapore},
doi = {10.1109/ICRA.2017.7989381}
}

@article{loquercio2018dronet,
author={Antonio Loquercio and Ana Isabel Maqueda and Carlos R. Del Blanco and Davide Scaramuzza},
title={{DroNet}: Learning to Fly by Driving},
journal={IEEE Robotics and Automation Letters},
year={2018},
volume={3},
number={2},
pages={1088--1095},
doi={10.1109/LRA.2018.2795643}
}

@article{mero2022adsurvey,
author={Le Mero, Luc and Yi, Dewei and Dianati, Mehrdad and Mouzakitis, Alexandros},
title={A Survey on Imitation Learning Techniques for End-to-End Autonomous Vehicles}, 
journal={IEEE Transactions on Intelligent Transportation Systems}, 
year={2022},
volume={23},
number={9},
pages={14128-14147},
doi={10.1109/TITS.2022.3144867}
}

@inproceedings{liao2025diffusiondrive,
author={Bencheng Liao and Shaoyu Chen and Haoran Yin and Bo Jiang and Cheng Wang and Sixu Yan and Xinbang Zhang and Xiangyu Li and Ying Zhang and Qian Zhang and Xinggang Wang},
title={{DiffusionDrive}: Truncated Diffusion Model for End-to-End Autonomous Driving},
booktitle={IEEE/CVF Conference on Computer Vision and Pattern Recognition (CVPR)}, 
year={2025},
address={Nashville, TN, USA},
doi={10.1109/CVPR52734.2025.01124}
}

@misc{nvidia2026alpamayo,
title={Alpamayo-R1: Bridging Reasoning and Action Prediction for Generalizable Autonomous Driving in the Long Tail}, 
author={NVIDIA and others},
year={2026},
eprint={2511.00088},
archivePrefix={arXiv},
primaryClass={cs.RO},
url={https://arxiv.org/abs/2511.00088}
}

@inproceedings{fisher2020bdd100k,
author={Yu, Fisher and Chen, Haofeng and Wang, Xin and Xian, Wenqi and Chen, Yingying and Liu, Fangchen and Madhavan, Vashisht and Darrell, Trevor},
title={BDD100K: A Diverse Driving Dataset for Heterogeneous Multitask Learning}, 
booktitle={IEEE/CVF Conference on Computer Vision and Pattern Recognition (CVPR)}, 
year={2020},
pages={2633-2642},
doi={10.1109/CVPR42600.2020.00271}
}

@inproceedings{sun2020waymo,
author={Sun, Pei and others},
booktitle={IEEE/CVF Conference on Computer Vision and Pattern Recognition (CVPR)}, 
title={Scalability in Perception for Autonomous Driving: Waymo Open Dataset}, 
year={2020},
pages={2443-2451},
doi={10.1109/CVPR42600.2020.00252}
}

@inproceedings{shah2023vint,
title={Vi{NT}: A Foundation Model for Visual Navigation},
author={Dhruv Shah and Ajay Sridhar and Nitish Dashora and Kyle Stachowicz and Kevin Black and Noriaki Hirose and Sergey Levine},
booktitle={Conference on Robot Learning (CoRL)},
year={2023},
address={Atlanta, GA, USA}
}

@inproceedings{sridhar2024nomad,
author= {Ajay Sridhar and Dhruv Shah and Catherine Glossop and Sergey Levine},
title= {{NoMaD: Goal Masked Diffusion Policies for Navigation and Exploration}},
booktitle={IEEE International Conference on Robotics and Automation (ICRA)}, 
year= {2024},
doi= {10.1109/ICRA57147.2024.10610665},
address={Yokohama, Japan}
}

@inproceedings{gode2025flownav,
author={Gode, Samiran and Nayak, Abhijeet and Oliveira, Débora N.P. and Krawez, Michael and Schmid, Cordelia and Burgard, Wolfram},
title={FlowNav: Combining Flow Matching and Depth Priors for Efficient Navigation}, 
booktitle={IEEE/RSJ International Conference on Intelligent Robots and Systems (IROS)},
year={2025},
address={Hangzhou, Zhejiang, China},
doi={10.1109/IROS60139.2025.11246634}
}

@inproceedings{shah2022viking, 
author={Dhruv Shah and Sergey Levine}, 
title={{ViKiNG: Vision-Based Kilometer-Scale Navigation with Geographic Hints}}, 
booktitle={Proceedings of Robotics: Science and Systems (RSS)}, 
year={2022},
address={New York, NY, USA},
doi={10.15607/RSS.2022.XVIII.019}
}

@inproceedings{kahn2018coryhall,
author={Kahn, Gregory and Villaflor, Adam and Ding, Bosen and Abbeel, Pieter and Levine, Sergey},
title={Self-Supervised Deep Reinforcement Learning with Generalized Computation Graphs for Robot Navigation},
booktitle={IEEE International Conference on Robotics and Automation (ICRA)},
year={2018},
address={Brisbane, Australia},
doi={10.1109/ICRA.2018.8460655}
}

@article{hirose2019gostanford,
author={Hirose, Noriaki and Xia, Fei and Martín-Martín, Roberto and Sadeghian, Amir and Savarese, Silvio},
journal={IEEE Robotics and Automation Letters}, 
title={Deep Visual MPC-Policy Learning for Navigation}, 
year={2019},
volume={4},
number={4},
pages={3184-3191},
doi={10.1109/LRA.2019.2925731}
}

@misc{agha2021nebula,
title={{NeBula: Quest for Robotic Autonomy in Challenging Environments; TEAM CoSTAR at the DARPA Subterranean Challenge}}, 
author={Ali Agha and others},
year={2021},
eprint={2103.11470},
archivePrefix={arXiv},
primaryClass={cs.RO},
url={https://arxiv.org/abs/2103.11470}
}

@inproceedings{shah2021recon,
author={Shah, Dhruv and Eysenbach, Benjamin and Rhinehart, Nicholas and Levine, Sergey},
title={Rapid Exploration for Open-World Navigation with Latent Goal Models},
booktitle={Conference on Robot Learning (CoRL)},
year={2021},
address={London, UK}
}

@article{hirose2024sacson,
author={Hirose, Noriaki and Shah, Dhruv and Sridhar, Ajay and Levine, Sergey},
title={{SACSoN: Scalable Autonomous Control for Social Navigation}}, 
journal={IEEE Robotics and Automation Letters}, 
year={2024},
volume={9},
number={1},
pages={49-56},
doi={10.1109/LRA.2023.3329626}
}

@article{karnan2022scand,
author={Karnan, Haresh and Nair, Anirudh and Xiao, Xuesu and Warnell, Garrett and Pirk, S{\"o}ren and Toshev, Alexander and Hart, Justin and Biswas, Joydeep and Stone, Peter},
title={{Socially CompliAnt Navigation Dataset (SCAND): A Large-Scale Dataset Of Demonstrations For Social Navigation}},
journal={IEEE Robotics and Automation Letters},
year={2022},
doi={10.1109/LRA.2022.3184025}
}

@inproceedings{shaban2021seattle,
author={Shaban, Amirreza and Meng, Xiangyun and Lee, JoonHo and Boots, Byron and Fox, Dieter},
title={Semantic Terrain Classification for Off-Road Autonomous Driving},
booktitle={Conference on Robot Learning (CoRL)},
address={London, UK},
year={2021}
}

@inproceedings{triest2022tartandrive,
author={Triest, Samuel and Sivaprakasam, Matthew and Wang, Sean J. and Wang, Wenshan and Johnson, Aaron M. and Scherer, Sebastian},
title={TartanDrive: A Large-Scale Dataset for Learning Off-Road Dynamics Models},
booktitle={IEEE International Conference on Robotics and Automation (ICRA)},
year={2022},
address={Philadelphia, PA, USA},
doi={10.1109/ICRA46639.2022.9811648}
}

@inproceedings{hirose2024lelan,
author={Noriaki Hirose and Catherine Glossop and Ajay Sridhar and Dhruv Shah and Oier Mees and Sergey Levine},
title={LeLaN: Learning A Language-conditioned Navigation Policy from In-the-Wild Video},
booktitle={Conference on Robot Learning (CoRL)},
year={2024},
address={Munich, Germany}
}

@inproceedings{liu2025citywalker,
author={Liu, Xinhao and Li, Jintong and Jiang, Yicheng and Sujay, Niranjan and Yang, Zhicheng and Zhang, Juexiao and Abanes, John and Zhang, Jing and Feng, Chen},
title={Citywalker: Learning embodied urban navigation from web-scale videos},
booktitle={IEEE/CVF Conference on Computer Vision and Pattern Recognition (CVPR)}, 
year={2025},
address={Nashville, TN, USA},
doi={10.1109/CVPR52734.2025.00645}
}

@article{hirose2026driveanywhere,
author={Hirose, Noriaki and Ignatova, Lydia and Stachowicz, Kyle and Glossop, Catherine and Levine, Sergey and Shah, Dhruv},
title={Learning to Drive Anywhere With Model-Based Reannotation}, 
journal={IEEE Robotics and Automation Letters}, 
year={2026},
volume={11},
number={2},
pages={1242-1249},
doi={10.1109/LRA.2025.3640424}
}

@misc{hirose2025omnivla,
author={Noriaki Hirose and Catherine Glossop and Dhruv Shah and Sergey Levine},
title={{OmniVLA: An Omni-Modal Vision-Language-Action Model for Robot Navigation}}, 
year={2025},
eprint={2509.19480},
archivePrefix={arXiv},
primaryClass={cs.RO},
url={https://arxiv.org/abs/2509.19480}
}

@inproceedings{patel2025tartanground,
author={Patel, Manthan and Yang, Fan and Qiu, Yuheng and Cadena, Cesar and Scherer, Sebastian and Hutter, Marco and Wang, Wenshan},
title={{TartanGround: A Large-Scale Dataset for Ground Robot Perception and Navigation}}, 
booktitle={IEEE/RSJ International Conference on Intelligent Robots and Systems (IROS)},
year={2025},
address={Hangzhou, Zhejiang, China},
doi={10.1109/IROS60139.2025.11246002}
}

@inproceedings{stachowicz2024liren,
author={Kyle Stachowicz and Lydia Ignatova and Sergey Levine},
title={Lifelong Autonomous Improvement of Navigation Foundation Models in the Wild},
booktitle={Conference on Robot Learning (CoRL)},
year={2024},
address={Munich, Germany}
}

@inproceedings{cheng2025navila,
author={Cheng, An-Chieh and Ji, Yandong and Yang, Zhaojing and Zou, Xueyan and Kautz, Jan and Biyik, Erdem and Yin, Hongxu and Liu, Sifei and Wang, Xiaolong},
title={{NaVILA: Legged Robot Vision-Language-Action Model for Navigation}},
booktitle={Proceedings of Robotics: Science and Systems (RSS)},
year={2025},
address={Los Angeles, CA, USA}
}

@misc{zhang2025ventura,
author={Arthur Zhang and Xiangyun Meng and Luca Calliari and Dong-Ki Kim and Shayegan Omidshafiei and Joydeep Biswas and Ali Agha and Amirreza Shaban},
title={{VENTURA: Adapting Image Diffusion Models for Unified Task Conditioned Navigation}}, 
year={2025},
eprint={2510.01388},
archivePrefix={arXiv},
primaryClass={cs.RO},
url={https://arxiv.org/abs/2510.01388}
}

@misc{peng2025logoplanner,
author={Jiaqi Peng and Wenzhe Cai and Yuqiang Yang and Tai Wang and Yuan Shen and Jiangmiao Pang},
title={LoGoPlanner: Localization Grounded Navigation Policy with Metric-aware Visual Geometry}, 
year={2025},
eprint={2512.19629},
archivePrefix={arXiv},
primaryClass={cs.RO},
url={https://arxiv.org/abs/2512.19629}
}

@inproceedings{kim2024openvla,
author={Kim, Moo Jin and others},
title={Open{VLA}: An Open-Source Vision-Language-Action Model},
booktitle={Conference on Robot Learning (CoRL)},
year={2024},
address={Munich, Germany}
}

@inproceedings{xu2017bddv,
author={Huazhe Xu and Yang Gao and Fisher Yu and Trevor Darrell},
title= {End-to-End Learning of Driving Models from Large-Scale Video Datasets},
booktitle={IEEE Conference on Computer Vision and Pattern Recognition (CVPR)}, 
year={2017},
address={Honolulu, HI, USA},
doi={10.1109/CVPR.2017.376}
}

@inproceedings{juan2025bodymppi,
title={Real-time whole-body control of legged robots with model-predictive path integral control},
author={Alvarez-Padilla, Juan and Zhang, John Z and Kwok, Sofia and Dolan, John M and Manchester, Zachary},
booktitle={IEEE International Conference on Robotics and Automation (ICRA)},
year={2025},
address={Atlanta, GA, USA},
doi={10.1109/ICRA55743.2025.11128271}
}

@inproceedings{miki2022elevation,
title={Elevation mapping for locomotion and navigation using gpu},
author={Miki, Takahiro and Wellhausen, Lorenz and Grandia, Ruben and Jenelten, Fabian and Homberger, Timon and Hutter, Marco},
booktitle={IEEE/RSJ International Conference on Intelligent Robots and Systems (IROS)},
year={2022},
address = {Kyoto, Japan},
doi={10.1109/IROS47612.2022.9981507}
}

@inproceedings{erni2023mem,
title={{MEM: Multi-Modal Elevation Mapping for Robotics and Learning}},
author={Erni, Gian and Frey, Jonas and Miki, Takahiro and Mattamala, Matias and Hutter, Marco},
booktitle={IEEE/RSJ International Conference on Intelligent Robots and Systems (IROS)},
year={2023},
address={Detroit, Michigan, USA},
doi={10.1109/IROS55552.2023.10342108}
}

@article{asad2022fastgeodis, 
author={Muhammad Asad and Reuben Dorent and Tom Vercauteren}, 
title={{FastGeodis: Fast Generalised Geodesic Distance Transform}}, 
journal={Journal of Open Source Software},
year={2022}, 
publisher={The Open Journal}, 
volume={7}, 
number={79}, 
pages={4532}, 
doi={10.21105/joss.04532}
}

@article{oquab2024dinov,
title={{DINO}v2: Learning Robust Visual Features without Supervision},
author={Maxime Oquab and others},
journal={Transactions on Machine Learning Research},
issn={2835-8856},
year={2024}
}

@inproceedings{loshchilov2019adamw,
author={Ilya Loshchilov and Frank Hutter},
title={Decoupled Weight Decay Regularization},
booktitle={International Conference on Learning Representations (ICLR)},
year={2019},
address={New Orleans, LA, USA}
}

@inproceedings{smith2019onecycle, 
author={Smith, Leslie N. and Topin, Nicholay}, 
title={Super-convergence: very fast training of neural networks using large learning rates}, 
booktitle={Artificial Intelligence and Machine Learning for Multi-Domain Operations Applications}, 
publisher={SPIE}, 
pages={36},
year={2019}, 
doi={10.1117/12.2520589}
}

@inproceedings{wang2025vggt,
author={Wang, Jianyuan and Chen, Minghao and Karaev, Nikita and Vedaldi, Andrea and Rupprecht, Christian and Novotny, David},
title={{VGGT}: Visual Geometry Grounded Transformer},
booktitle={IEEE/CVF Conference on Computer Vision and Pattern Recognition (CVPR)}, 
year={2025},
address={Nashville, TN, USA},
doi={10.1109/CVPR52734.2025.00499}
}

\clearpage 
\nobalance

\appendix
\label[appsec]{appendix}

\subsection{Implementation Details}
\label[appsec]{app:implementation-detail}

\subsubsection{Teleoperation Dataset (\Dtel)}
\label[appsec]{app:tel-curation-details}
We extract goal-conditioned training samples from \Dbase, which provides time-synchronized images and state estimates from teleoperated missions.
Let $\mathbf{\bar{s}}(t)\in\SeTwo$ denote the recorded robot pose at time $t$.
For each camera frame at time $t_0$ with image $\mathbf{I}(t_0)$, we sample a future time $t_g>t_0$ and set the goal to the corresponding future pose, expressed in the robot-centric frame at $t_0$.

We construct a fixed-horizon waypoint sequence of length $N$ with step size $\Delta t$ (horizon $T=N\Delta t$) by sampling poses from the teleoperation trajectory.
We sample $N$ equally spaced timestamps over the interval $[t_0,\,t_0+\min(T,\,t_g-t_0)]$ and set each waypoint to the recorded pose at that timestamp, \ie, $\mathbf{s}_i=\mathbf{\bar{s}}(t_i)$.
If $t_g-t_0>T$, the segment covers the next $T$ seconds of teleoperation and does not extend to the goal.

If $t_g-t_0\le T$, we time-warp the segment by sampling the full interval $[t_0,t_g]$ into $N$ waypoints, ensuring a fixed output horizon and smoothing abrupt decelerations near the goal.

Each sample is stored as $(\mathbf{I}, \mathbf{g}, \mathbf{s}_{1:N})$ in \Dtel.

\subsubsection{Traversability Mapping}
The CNN output is thresholded into discrete regions.
Cells with $<0.3$ are classified as safe and assigned a traversability cost of $0$. 
Between $0.3$ and $0.8$, we use a linear ramp to map to corresponding traversability costs. 
Cells above $0.8$ are deemed risky and assigned a cost of $2.0$. 
Anything above $0.9$ is an obstacle and is assigned a cost of $10^5$.

\subsubsection{Goal Sampling}
For each frame, we draw $K$ goals from
$\!\mathcal N\!\big([\,\SI{5}{m},\,\SI{0}{m},\,0\,]^\top,\operatorname{diag}(({2.5}~\text{m})^2,\,(2.0~\text{m})^2,\!(\pi/4)^2)\big)$.

\subsubsection{Rollouts and Footprint}
MPPI uses a horizon of $N=50$ with $\Delta t=\SI{0.1}{s}$. The robot footprint (dilated by a safety margin) is rasterized on the grid for collision queries along rollouts.

\subsubsection{Dataset Filtering}
We discard stationary or near-stationary segments with a final displacement $<\!\SI{0.25}{m}$ to reduce redundancy in \Dtel.

In \Dgeo, we discard frames in which the elevation map contains fewer than 25\% valid measurements.

\subsubsection{Gridmaps}
The traversability and geodesic distance fields are represented as $8 \times 8$~m grid maps with a $4 \times 4$~cm resolution, centered around the robot. 

\subsubsection{Architecture}
Our policy consists of a frozen DINOv2 ViT-S/14~\cite{oquab2024dinov} image encoder and a transformer decoder with four blocks for waypoint prediction.
All embeddings (tokens) have dimension $384$, matching the embedding dimension of DINOv2 ViT-S/14.
All DINOv2 parameters are frozen except for LayerNorm layers.
The full model contains approximately \SI{32}{M} parameters, of which \SI{10}{M} are trainable.

\subsubsection{Training}
We train \limo using standard supervised regression with an L2-loss on the predicted waypoints and train for 16~epochs using AdamW~\cite{loshchilov2019adamw} with a OneCycleLR~\cite{smith2019onecycle} schedule.
This corresponds to \SI{1.4}{M} steps with a batch size of 16 on a single NVIDIA RTX-4090 GPU.
Longer training did not further improve performance. 

\subsection{\emph{OmniVLA} Comparison}
\label[appsec]{app:omnivla-comparison}

OmniVLA~\cite{hirose2025omnivla} is a visual navigation model that supports multiple goal-conditioning modalities (\SeTwo goal pose, goal image, language instructions, and their combinations) and predicts short-horizon navigation commands.
In the released inference code, OmniVLA outputs a sequence of $8$ future ``actions,'' corresponding to \emph{relative waypoints} (consistent with the ViNT~\cite{shah2023vint} and NoMaD~\cite{sridhar2024nomad} representation) that are tracked by a low-level controller.

OmniVLA is trained to imitate a fixed-length chunk of $8$ waypoints, even when the provided goal lies within this horizon.
According to the authors (email communication), the goal pose is treated as one of several subgoals along a trajectory, and the model is trained to predict the full waypoint chunk, regardless of whether the goal is reached before step~8.

As a result, OmniVLA is not explicitly trained to produce trajectories that terminate at the goal in open-loop; instead, stopping behavior is handled by the downstream waypoint tracker.

\paragraph{Evaluation Protocol}
We evaluate OmniVLA in pose-only conditioning mode using the released checkpoints.
To enable a fair comparison in open-loop evaluation, we report success if the \emph{closest} waypoint in the predicted sequence is within \SI{1}{m} of the goal, rather than requiring the final waypoint to coincide with it.
Predicted waypoints are de-normalized using the scaling constant (\texttt{metric\_waypoint\_spacing}) as in the official code, chosen to match the scale of our local planning horizon.

\paragraph{Model Scale and Motion Diversity}
OmniVLA's support of broad conditioning modalities comes at the cost of high model complexity and size.
It builds on OpenVLA~\cite{kim2024openvla}, a \SI{7}{B}-parameter vision-language-action model pretrained primarily for manipulation. 
OmniVLA uses pretraining from OpenVLA and is further adapted for navigation using a cross-embodiment dataset heavily dominated by wheeled platforms (\eg, BDD-V~\cite{xu2017bddv}, Frodobots-2K~\cite{hirose2026driveanywhere}).
Consequently, OmniVLA produces paths that are predominantly forward-moving, and conditioning signals predominantly modulate steering rather than enabling diverse motion behaviors.

In contrast, \limo is a lightweight model with \SI{32}{M} parameters (\SI{10}{M} trainable), trained for embodiment-specific navigation tasks.
It predicts $50$ waypoints, providing finer temporal resolution and longer horizon planning.
Its training data includes sideways, backward, and in-place motions characteristic of legged locomotion, enabling full \SeTwo planning consistent with quadruped behavior.

\paragraph{Discussion}
Under this protocol, OmniVLA exhibits competitive collision avoidance but substantially lower goal-reaching performance than \limo on our evaluation (\cref{tab:quant-results}).
We attribute this primarily to OmniVLA’s closed-loop training assumptions and the strong bias toward wheeled-platform motion in its training data.

Moreover, OmniVLA is trained as a cross-embodiment policy without explicit knowledge of the target robot’s physical characteristics and therefore lacks an internal notion of the platform's size and maneuverability.
This absence of embodiment awareness hinders precise goal-reaching, particularly for a legged robot whose motion capabilities differ substantially from the car-like behaviors.

In contrast, \limo's training is explicitly tailored to ANYmal’s geometry and traversability model, enabling more accurate and platform-consistent planning.

Overall, this comparison suggests that simply scaling navigation policies to large, heterogeneous datasets does not necessarily yield robust, accurate, goal-conditioned planning.  
This raises the broader question of whether embodiment-agnostic foundation models suffice for robotic navigation, or whether effective navigation requires supervision grounded in the geometry and constraints of the target robot.

\subsection{Incorporating Side Cameras}
\label[appsec]{app:side-cams}

When relying solely on the front-facing camera, \limo cannot maintain awareness of obstacles once they leave the field of view.
This can lead to collisions when the policy corrects the robot's path before fully clearing an obstacle. 

Incorporating explicit memory or leveraging past observations could mitigate this issue, but we consider such extensions orthogonal to the focus of this work and leave them for future research.
Instead, we addressed the issue by adding the robot's left- and right-facing camera views to the training data and by minimally modifying the policy architecture to also use the side cameras. 

Image features are extracted using the same shared DINOv2~\cite{oquab2024dinov} encoder from all three camera views. 
The features are enriched with an additional learned absolute positional embedding, allowing the decoder to differentiate between features from different camera views. 
All image features are forwarded to the decoder; otherwise, it remains unchanged. 
This enables \limo to retain awareness of lateral obstacles (see~\cref{fig:side-view-model}).

This limitation is less critical for cross-embodiment models such as OmniVLA, whose training on predominantly wheeled platforms induces conservative, Ackermann-like behavior.
Because these models mainly adjust steering while moving forward, they rarely make early lateral corrections that provoke collisions when obstacles exit the forward field of view.

\limo, in contrast, deliberately exploits more agile motion capabilities of quadruped robots, making it more sensitive to this effect and motivating the use of side-camera observations.

\subsection{Limitations}
\label[appsec]{app:limitations}
Our method inherits limitations from the geometric supervisor.  
The geometric planning depends on elevation and traversability estimates that can fail in edge cases; for example, cliff edges may be classified as safe (\cref{fig:emergent}-d), and difficult terrain can lead to suboptimal geometric paths.

While expert demonstrations provide semantic cues, they are limited in scale, so \limo may still struggle when semantics play a critical role.
Also, simply combining the datasets on a per-sample basis might not be the optimal way to convey semantic and geometric cues. 
A mechanism for selectively choosing or weighting samples during training could be promising. 

We can scale the amount of supervision, but the visual diversity of the dataset's scenes remains fixed.  
Pretrained encoders, such as DINOv2~\cite{oquab2024dinov}, help; yet out-of-distribution cases, for example, the net and snow in \cref{fig:emergent}-m, remain challenging.

Finally, \limo operates on a single RGB frame without temporal memory, which limits its ability to handle occlusions or reason over longer time horizons.
This can lead to \limo forgetting obstacles when they leave the field of view, or to oscillations or temporally inconsistent plans.

\subsection{Future Work}
\label[appsec]{app:future-work}

Several extensions could further improve performance. 

(i) Following Alpamayo-R1~\cite{nvidia2026alpamayo}, direct waypoint regression could be replaced with a discretized trajectory representation and a generative objective such as conditional flow matching, which may better capture the inherently multi-modal distribution of feasible paths in ambiguous scenes.

(ii) Our geometric supervision currently relies on co-registered elevation maps; leveraging camera-only datasets by extracting scene geometry using pretrained models such as VGGT~\cite{wang2025vggt} could broaden the training distribution.

(iii) More principled strategies for sampling or weighting expert versus planner supervision may improve the transfer of semantic preferences while preserving strong geometric safety.

(iv) Incorporating temporal context is another promising direction: memory-based approaches could maintain awareness of previously observed obstacles without requiring additional cameras.

\ifanonymous
\else 
\section*{Acknowledgments}
The authors would like to thank Turcan Tuna and William Talbot for their support during robot experiments and system integration.
We also thank members of the Robotic Systems Lab at ETH Zurich for valuable discussions and feedback throughout the project.
We thank the authors of OmniVLA for providing model checkpoints, code snippets, and helpful clarifications.
\fi

\clearpage

\begin{table*}[t]

\begin{minipage}{0.48\linewidth}
\includegraphics[width=\linewidth]{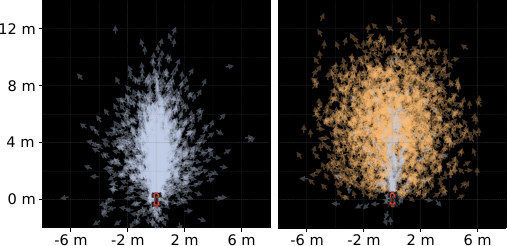}

\refstepcounter{figure}
\label{fig:goals}
\vspace{0.5em}

\setlength{\parindent}{0pt}
Fig.~\thefigure. Visualization of goal distributions based on 10,000 samples from each dataset.
(Left) Goals from the teleoperation dataset \Dtel (blue), sampled along the recorded trajectories.
(Right) Goals from the augmented dataset \Daug, including teleoperation goals (blue) and additional randomly sampled goals (orange).

\end{minipage}
\hfill
\begin{minipage}{0.48\linewidth}
\centering
\caption{Dataset statistics}
\label{tab:dataset_stats}

\begin{tabular}{@{}l l
                S[table-format=7]
                S[table-format=7]
                S[table-format=4]
                S[table-format=1.2]@{}}
    \toprule
    {Dataset} & {Split} & {\#Samples} & {Length [m]} & {Time [h]} & {Avg.\ vel. [m/s]} \\
    \midrule
    \Dtel & Train &  113326 &  281870 &  157 & 0.50 \\
         & Test  &   17139 &    43212 &   24 & 0.50 \\
    \addlinespace
    \Dgeo & Train & 1289930 & 4457729 & 1792 & 0.69 \\
         & Test  &  191060 &   653978 &  265 & 0.68 \\
    \addlinespace
    \Daug & Train & 1403256 & 4739599 & 1949 & 0.68 \\
         & Test  &  208199 &   697190 &  289 & 0.67 \\
    \bottomrule
\end{tabular}

\vspace{0.5em}
\parbox{\linewidth}{
Each sample is a fixed-horizon waypoint sequence with duration $T=\SI{5}{s}$.
The reported path length is the sum of Euclidean distances between consecutive waypoints.
The reported average velocity should be interpreted with care: while the robot typically walks at around \SI{1}{m/s}, many samples include stopping behavior (\eg, trajectories that reach the goal early), which lowers the mean speed.
}
\end{minipage}

\end{table*}

\begin{figure*}
    \centering
    \includegraphics[width=1.0\linewidth]{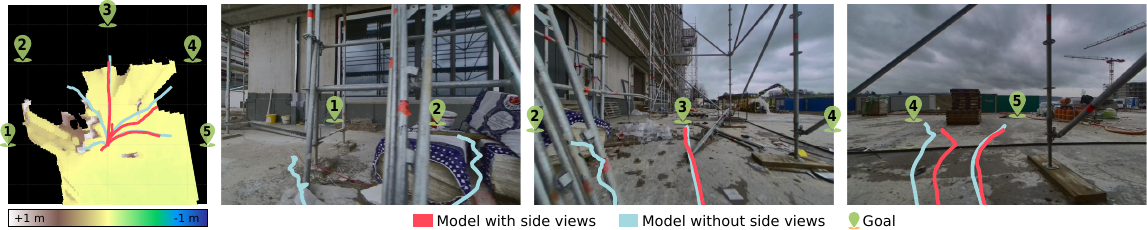}
    \caption{
    Comparison of trajectory predictions with and without side-facing cameras.
    Five goals (green) and the corresponding predicted trajectories are shown.
    Using only the front camera (cyan), \limo loses awareness of a partially cleared obstacle once it leaves the forward field of view, leading to collisions.
    With left and right side camera views (red), obstacle awareness is preserved, and all paths remain collision-free.
    }
    \label{fig:side-view-model}
\end{figure*}

\begin{figure*}
    \centering
    \includegraphics[width=1.0\linewidth]{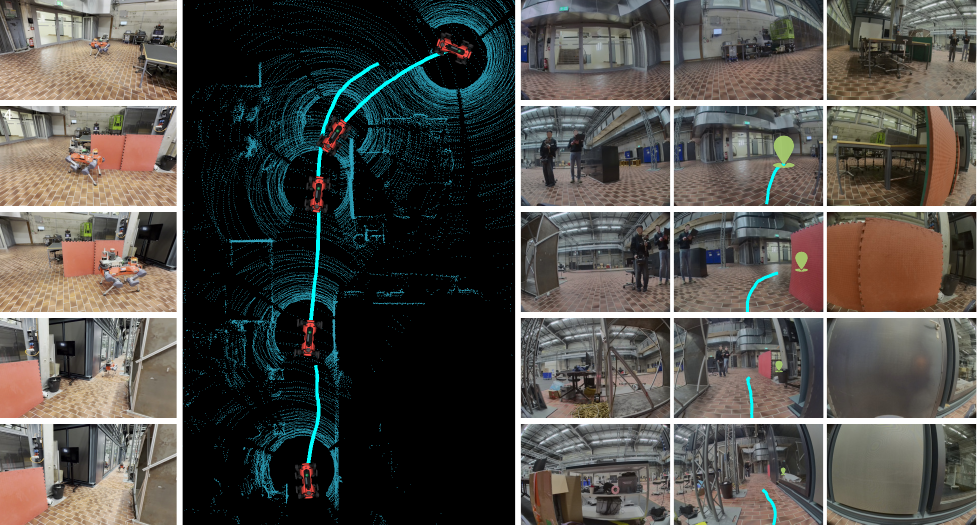}
    \caption{
    \limo is following a narrow corridor toward a goal located at the far end.
    From left to right: (i) third-person view, (ii) bird’s-eye view of the LiDAR point cloud with predicted paths, and (iii) left-, front-, and right-facing camera views.
    The predicted trajectory (cyan) remains centered and collision-free despite tight lateral constraints.
    }
    \label{fig:corridor}
\end{figure*}

\end{document}